\documentclass[12pt]{article}

\usepackage[english]{babel}
\usepackage{sbc-template}
\usepackage{graphicx,url}
\usepackage[utf8]{inputenc}
\usepackage[english]{babel}
\usepackage{hyperref}

\usepackage{tabularx}
\usepackage{booktabs}
\usepackage{array}
\usepackage{comment}
\newcolumntype{Y}{>{\raggedright\arraybackslash}X}
     
\title{Class Imbalance and Batch Effects in LLM-Based Screening for Systematic Reviews}

\author{Gilberto Sussumu Hida\inst{1}, 
Danilo Monteiro Ribeiro\inst{1}, Clayton Suguio Hida\inst{2}}

\address{Cesar School -- AIBL\\
Recife - PE - Brazil
\nextinstitute
UEAP\\
Macapá - AP - Brazil
\email{gsh@cesar.school, dmr@cesar.school, clayton.hida@ueap.edu.br}
}
\begin{document} 

\maketitle
\begin{center}
\footnotesize\textit{This is the authors' version of the paper accepted for publication at ENIAC 2026 (National Meeting on Artificial and Computational Intelligence, part of BRACIS 2026). The definitive version will be published in the conference proceedings by the Brazilian Computer Society (SBC).}
\end{center}
\begin{abstract}
This study analyses LLMs in imbalanced binary classification, using study screening in systematic reviews as the application domain. An experiment was conducted in five reviews, comparing individual and batch processing, with and without prevalence metadata. The results indicate a limited influence of the prevalence metadata, with no evidence that it improves performance. In contrast, batch processing produced larger behavioral changes that varied according to the prevalence of the class. The aggregate and item-level analyses did not always coincide. Therefore, batch processing should be evaluated not only in terms of cost, but also in relation to its effects on decision-making behavior.
\end{abstract}

\section{Introduction}
Systematic literature reviews (SLRs) have become a central practice in Software Engineering~\cite{kitchenham2004procedures}. With the growth in publication volume~\cite{bornmann2021growth}, the title and abstract screening stage has become increasingly costly, requiring the classification of hundreds to thousands of records~\cite{chai2021research}. Large language models (LLMs) have been explored as a support for this task~\cite{hida2026beyond,syriani2024screening,huotala2025sesr}, with promising results in aggregate performance, but still limited understanding of their decision behavior under operational screening characteristics.

Two of these characteristics are methodologically central. The first is class imbalance: in most screening processes, only a fraction of records is included, making traditional metrics misleading and shifting the optimal point toward minimizing false negatives~\cite{hida2025overview}. The second is the processing mode: LLMs can receive one study per call (individual) or multiple studies in a single call (batch), which reduces cost~\cite{pipal2026researchers} but can induce implicit comparison between candidates, effects of quota or sensitivity to position in batch~\cite{liu2024lost}.

Recent literature has focused on comparisons between models and prompts, often under individual processing and without isolating the effect of contextual metadata~\cite{madeyski2025llm4screenlit,syriani2024screening}. It remains unclear whether informing screening metadata, such as expected prevalence, changes LLMs' propensity to include or exclude studies, and whether the processing mode has a comparable effect.

This work addresses this gap with a factorial design of 2$\times$2: presence or absence of screening metadata in the prompt and processing mode (individual or batch). We selected five reviews from SESR-Eval~\cite{huotala2025sesr}, with prevalence ranging from $2.9\%$ to $53.0\%$, and evaluated two models: Llama-3.3-70B-Instruct-Turbo~\cite{meta2024llama33} and GPT-5-mini~\cite{openai2025gpt5mini}.

The contributions are: (a) evidence that informing prevalence metadata in the prompt produced a limited effect, whereas batch processing has a larger and more variable behavioral effect across reviews; (b) the observation that aggregate and item-level analyses may diverge; and (c) implications for the responsible use of LLMs in screening, showing that how articles are presented to the model matters more than prompt-based calibration. 


\section{Related Work}

The application of LLMs to title and abstract screening has been investigated in systematic reviews, including in the Software Engineering domain. Syriani et al.~\cite{syriani2024screening} evaluated ChatGPT in this context and observed relevant variation between reviews. Huotala et al.~\cite{huotala2025sesr} proposed SESR-Eval, a benchmark with 34,528 labeled primary studies from 24 secondary reviews, showing that variation between reviews may exceed variation between models. In a methodological review, Madeyski and Kitchenham~\cite{madeyski2025llm4screenlit} highlight recurring limitations in the evaluation of LLMs for screening, including the underuse of metrics suitable for imbalanced scenarios.

Another line of work investigates how prompt formulation and calibration mechanisms influence LLM performance. Wang et al.~\cite{wang2024zero} evaluated LLMs in a zero-shot setting and proposed calibration based on target recall, while Huotala et al.~\cite{huotala2024promise} compared zero-shot, one-shot and few-shot configurations. These studies indicate that instructions, examples, and thresholds can alter performance, but they do not directly isolate the effect of contextual screening metadata, such as informed prevalence, on the models' decision propensity.

Batch processing has been treated mainly as an efficiency strategy. Cheng et al.~\cite{cheng2023batch} and Pipal et al.~\cite{pipal2026researchers} showed that batch prompting reduces token, time, and cost while maintaining performance close to individual processing. However, increasing the context may alter the model behavior. Liu et al.~\cite{liu2024lost} documented the Lost in the Middle phenomenon, in which performance varies according to the position of information in the context. Fagerberg et al.~\cite{fagerberg2026batch} showed that aggregate sensitivity for GPT-5-mini remains stable up to batch sizes of approximately 150–200 before degrading sharply. Thus, batch processing should be analyzed as an experimental condition, not only as a cost optimization.

Finally, evaluating LLMs in screening must consider class imbalance, since Include tends to be the minority class and accuracy-based metrics may be misleading. Madeyski and Kitchenham~\cite{madeyski2025llm4screenlit} discuss the divergences between traditional metrics and metrics that are more sensitive to asymmetric cost. Hida and Do Nascimento~\cite{hida2025overview} discuss robust metrics for imbalanced classification, while Hida et al.~\cite{hida2026beyond} show that variability between executions of the same LLM can be comparable to differences between models, recommending more stable agreement metrics and measures aligned with the cost of false negatives, such as the F2-score.

Although the literature advances in benchmarks, prompting, batch processing, and metrics, three gaps remain: batch processing is commonly treated as an efficiency technique, rather than as a behavioral factor; prompts rarely incorporate contextual screening metadata; and evaluation tends to emphasize aggregate metrics, without examining item-level changes in decisions. This study investigates these gaps through a controlled experimental design.


\section{Methodology}
\begin{figure}[ht]
    \centering
    \includegraphics[scale=0.28]{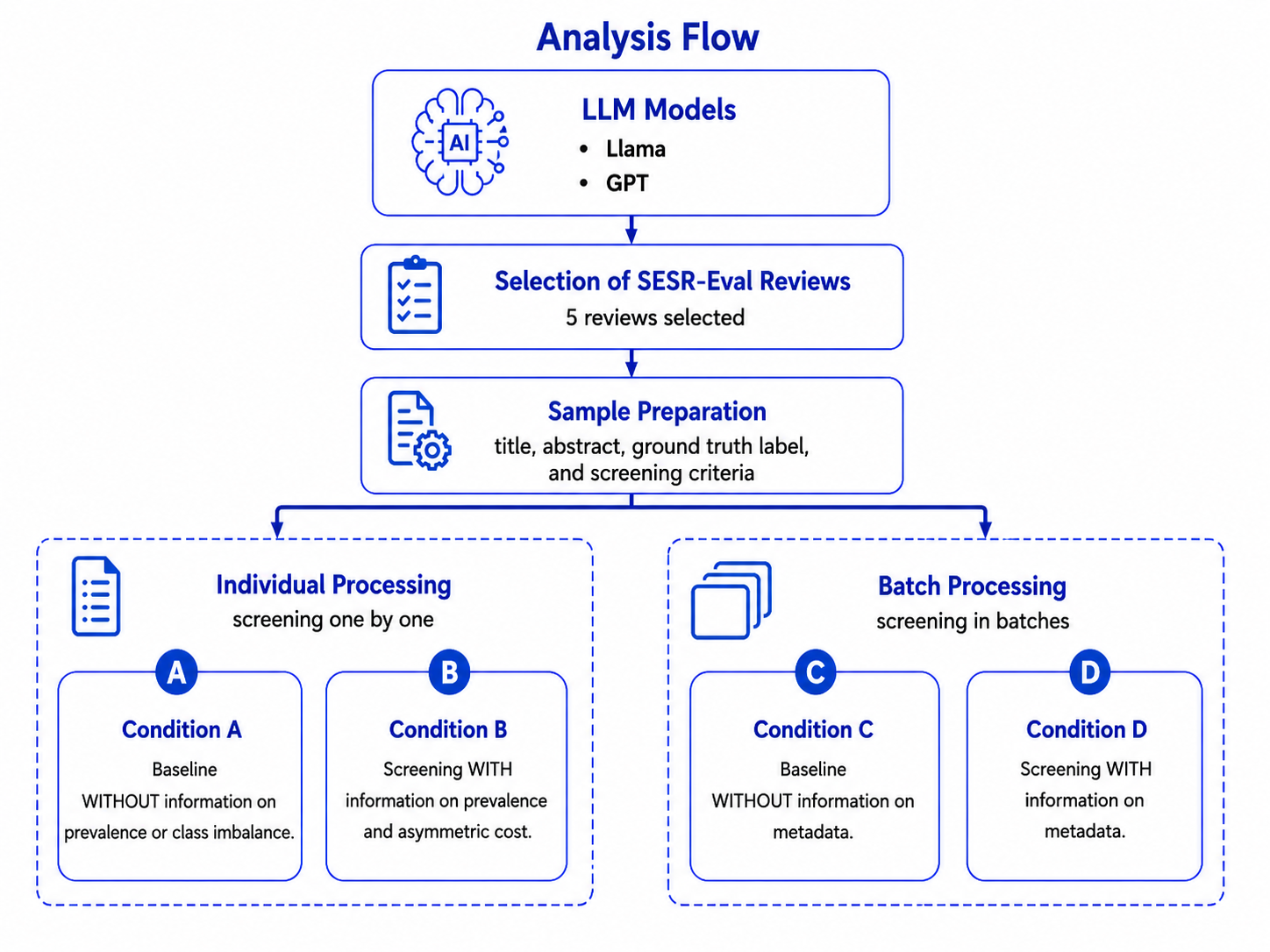}
    \caption{Workflow of the analysis conducted in the study.}
    \label{fig:fluxo_metodo}
\end{figure}

This controlled experiment evaluated how screening metadata and processing mode influence LLM behavior in a binary classification task involving scientific studies. Each candidate study was classified as \textit{Exclude} or \textit{Include}, based on its title, abstract, and eligibility criteria for the corresponding secondary review. Figure~\ref{fig:fluxo_metodo} summarizes the overall analysis workflow.

The analysis was guided by four research questions: 
\begin{itemize}
    \item \textbf{RQ1}) Do screening metadata change the propensity of LLMs to classify studies as \textit{Include}?
    \item \textbf{RQ2}) Does batch processing modify decisions compared to individual processing?
    \item \textbf{RQ3}) How does batch processing, with or without metadata, affect false negatives and direction of decision changes?
    \item \textbf{RQ4}) Are the observed effects consistent across models from different families?
\end{itemize}


\subsection{Experimental design}

The experimental design followed a factorial structure of 2$\times$2, which combined two factors: the presence or absence of metadata for the selection in the prompt and processing mode, individual or batch. The metadata consisted of a single sentence informing the true (oracle) estimated inclusion rate of the \textit{Include} class in the review, framed as class-imbalance context. The four conditions were: A) individual without metadata; B) individual with metadata; C) batch without metadata; and D) batch with metadata.

\subsection{Dataset, models, and data preparation}

We used SESR-Eval~\cite{huotala2025sesr}, a title-abstract screening dataset built from Software Engineering secondary studies. It was chosen because it provides reference labels and reviews spanning different Include-class prevalence levels, allowing observation of model behavior across degrees of imbalance.

Five secondary reviews were selected to cover different positive-class prevalence values, from highly imbalanced to almost balanced. Reviews lacking both classes, with few records, or with insufficient eligibility criteria were discarded. Table~\ref{tab:revisoes_selecionadas} presents the characterization of the reviews used.

\begin{table}[ht]
\centering
\caption{Characterization of the selected SESR-Eval reviews.}
\small
\begin{tabular}{llrrrr}
\hline
ID & Ref. & N & Inc. & Exc. & Prev. \\
\hline
rsl\_0 & \cite{kuutila2020time} & 1{,}812 & 53  & 1{,}759 & 2.9\% \\
rsl\_1 & \cite{rani2023decade} & 2{,}048 & 71  & 1{,}977 & 3.5\% \\
rsl\_2 & \cite{lewowski2022far} & 1{,}527 & 160 & 1{,}367 & 10.5\% \\
rsl\_3 & \cite{tambon2022certify} & 955  & 166 & 789  & 17.4\% \\
rsl\_4 & \cite{sharbaf2023conflict} & 300  & 159 & 141  & 53.0\% \\
\hline
\end{tabular}
\label{tab:revisoes_selecionadas}

\vspace{0.2em}
\footnotesize{\textit{Note:} N represents the number of studies considered after data preparation for analysis.}
\end{table}

The unit of analysis was the candidate study. Each instance contains the study identifier, title, abstract, reference label, and the secondary study to which it belongs. The original label was binarized as $0$ for \textit{Exclude} and $1$ for \textit{Include}, and the decisions made by the LLMs were treated with the same coding.

Two models were evaluated: Llama-3.3-70B-Instruct-Turbo~\cite{meta2024llama33}, accessed via the Together AI API~\cite{togetherai2024api}, and gpt-5-mini-2025-08-07~\cite{openai2025gpt5mini}, accessed via the OpenAI API. Experiments were executed between March and May 2026. The same elements, conditions, and prompt templates were used for both models. Calls were executed with temperature 0; invalid or unexpected-format responses were monitored, with no invalid decisions in the final set. The comparison between models was treated as a robustness analysis and not as a competitive ranking of performance.

\subsection{Prompts and scenario execution}

The prompts were written in English, following the language of the dataset titles, abstracts, and criteria (available in the artifact package \cite{anonymous_artifacts_2026}). All conditions followed a common structure that contained the task definition, eligibility criteria, screening instructions, and candidate-study data. The output was requested in JSON format, with the study identifier and binary decision, in order to reduce parsing ambiguities and enable programmatic validation. In batch conditions (C and D), studies were randomly shuffled with a fixed seed and grouped into batches of 20, with the same batch composition and order reused across conditions C and D; each batch returned one decision per study. This size remains well within the range shown to preserve stable LLM performance in screening tasks~\cite{fagerberg2026batch}. The exact prompt templates for all four conditions and both model providers are reproduced in the supplementary material accompanying the artifact package~\cite{anonymous_artifacts_2026}.

\subsection{Metrics, analytical comparisons, and uncertainty}

The F2-score was adopted because it assigns greater weight to \textit{recall}, which is appropriate for the screening context, in which false negatives are more critical than false positives~\cite{hida2026beyond}. The false negative rate represented the risk of losing relevant studies, while Gwet's AC1~\cite{vach2023gwet, hida2026beyond} was used as an agreement metric due to its stability in extreme-prevalence scenarios. Because universal interpretation thresholds for agreement coefficients are context-dependent, AC1 values were interpreted primarily in comparative terms across reviews, models, and experimental conditions, rather than as absolute quality categories. This choice is particularly relevant in imbalanced screening settings, where agreement coefficients may be affected by prevalence and marginal distributions.

In addition to aggregate metrics, item-level indicators were calculated. The Decision Flip Rate (DFR) was defined as the proportion of studies whose decision changed between two conditions, inspired by decision flip and prediction churn metrics~\cite{laban2023you}. The \textit{net shift} was defined as the net difference between the changes from \textit{Exclude} to \textit{Include} and from \textit{Include} to \textit{Exclude}, indicating whether a condition made the model more inclusive or more restrictive.

The comparisons A$\leftrightarrow$B and C$\leftrightarrow$D assessed the effect of metadata in the individual and batch modes, respectively; A$\leftrightarrow$C and B$\leftrightarrow$D assessed the effect of batch processing without and with metadata. McNemar's exact test and article-level improvement or worsening analysis were also used. Paired bootstrap~\cite{bestgen2022please} confidence intervals were calculated for differences in F2-score and DFR, preserving the correspondence between studies in the compared conditions. These procedures assume independence; batch decisions sharing a prompt may be correlated, so uncertainty in batch comparisons may be underestimated.

\subsection{Reproducibility}

The study artifacts, including data tables, Jupyter notebooks used to prepare the data, process the LLM outputs, and reconstruct the analyses, tables, and figures, and a supplementary document with the exact prompt templates for all four conditions, are available in a \href{https://doi.org/10.5281/zenodo.20414963}{Zenodo record} (DOI: 10.5281/zenodo.20414963).

\section{Results and Discussion}

This section examines how batch processing and screening metadata affect the decision behavior of LLMs. The analysis combines three sources of evidence: agreement with the reference label, changes at the item-level, and the aggregate impact on performance. 

\subsection{Overview of agreement}

\begin{figure}[ht]
    \centering
    \includegraphics[scale=.57]{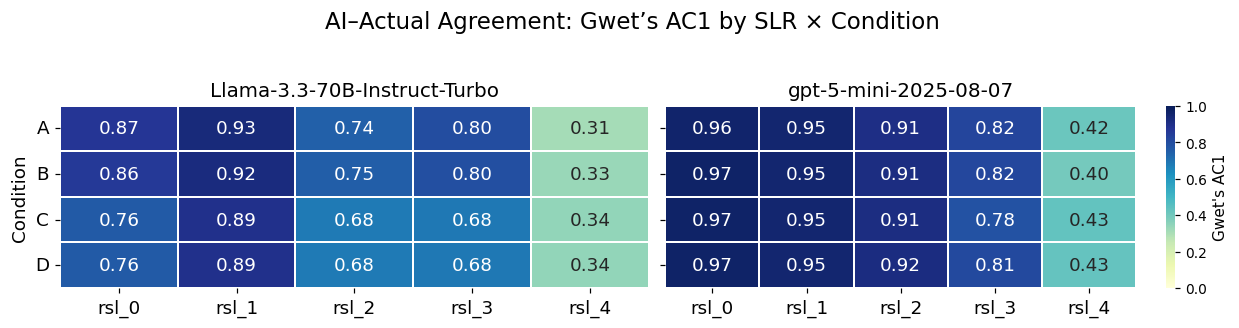}
    \caption{Gwet's AC1 by systematic review, experimental condition, and model.}
    \label{fig:heatmap_ac1}
\end{figure}

Figure~\ref{fig:heatmap_ac1} presents an overview of the agreement between LLM decisions and the reference label, using Gwet's AC1. The most evident pattern is that GPT-5-mini presented a higher and more stable agreement than Llama-3.3-70B-Instruct-Turbo in almost all reviews and conditions. In Llama, batch conditions reduced the agreement in reviews such as rsl\_0 and rsl\_3, suggesting that the mode of presentation of studies can alter the agreement between LLM decisions and the reference label. In GPT-5-mini, the variation between conditions was smaller, although the nearly balanced review (rsl\_4) also presented the lowest AC1 values.

This initial reading is important for two reasons. First, agreement varies more between reviews than between some conditions within the same review, indicating that the nature of the secondary study and the prevalence of the Include class influence the results. Second, given that the rsl\_i labels were ordered so that rsl\_0 has the lowest prevalence proportion ($<$3\%) and rsl\_4 the highest (53\%), the lower agreement in rsl\_4 shows that more balanced reviews are not necessarily easier for LLMs; in contrast, they may require finer distinctions between relevant and irrelevant studies.

\subsection{RQ1: Do screening metadata change the propensity of LLMs to classify studies as \textit{Include}?}

\begin{figure}[ht]
    \centering
    \includegraphics[scale=0.7]{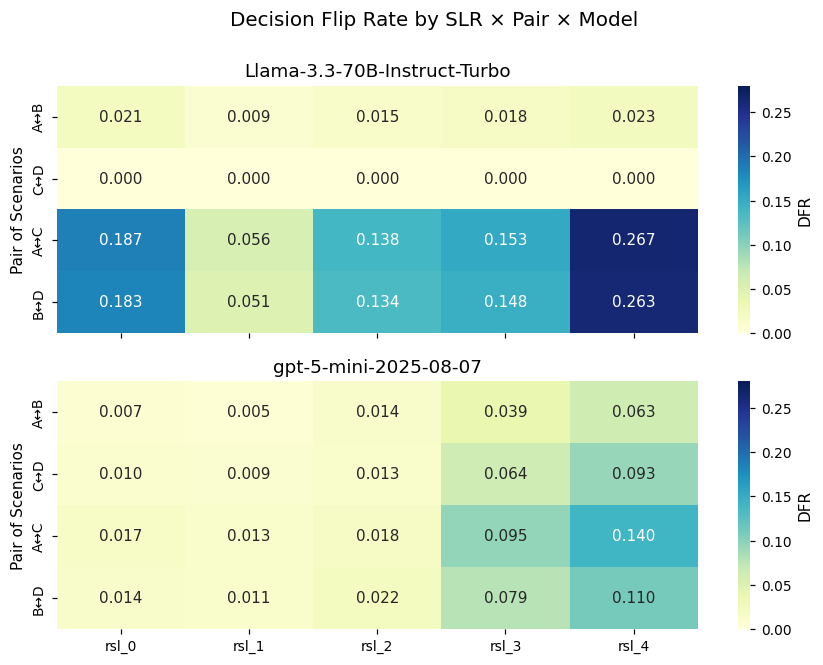}
    \caption{Decision Flip Rate by condition pair, SLR, and model.}
    \label{fig:RQ1_1}
\end{figure}

RQ1 examines whether the inclusion of screening metadata (informed prevalence) changes LLM decisions in two contexts: individual processing (A$\leftrightarrow$B) and batch processing (C$\leftrightarrow$D). Figure~\ref{fig:RQ1_1} presents the DFR for all pairwise comparisons, by model.

\begin{figure}[ht]
    \centering
    \includegraphics[width=1.0\linewidth]{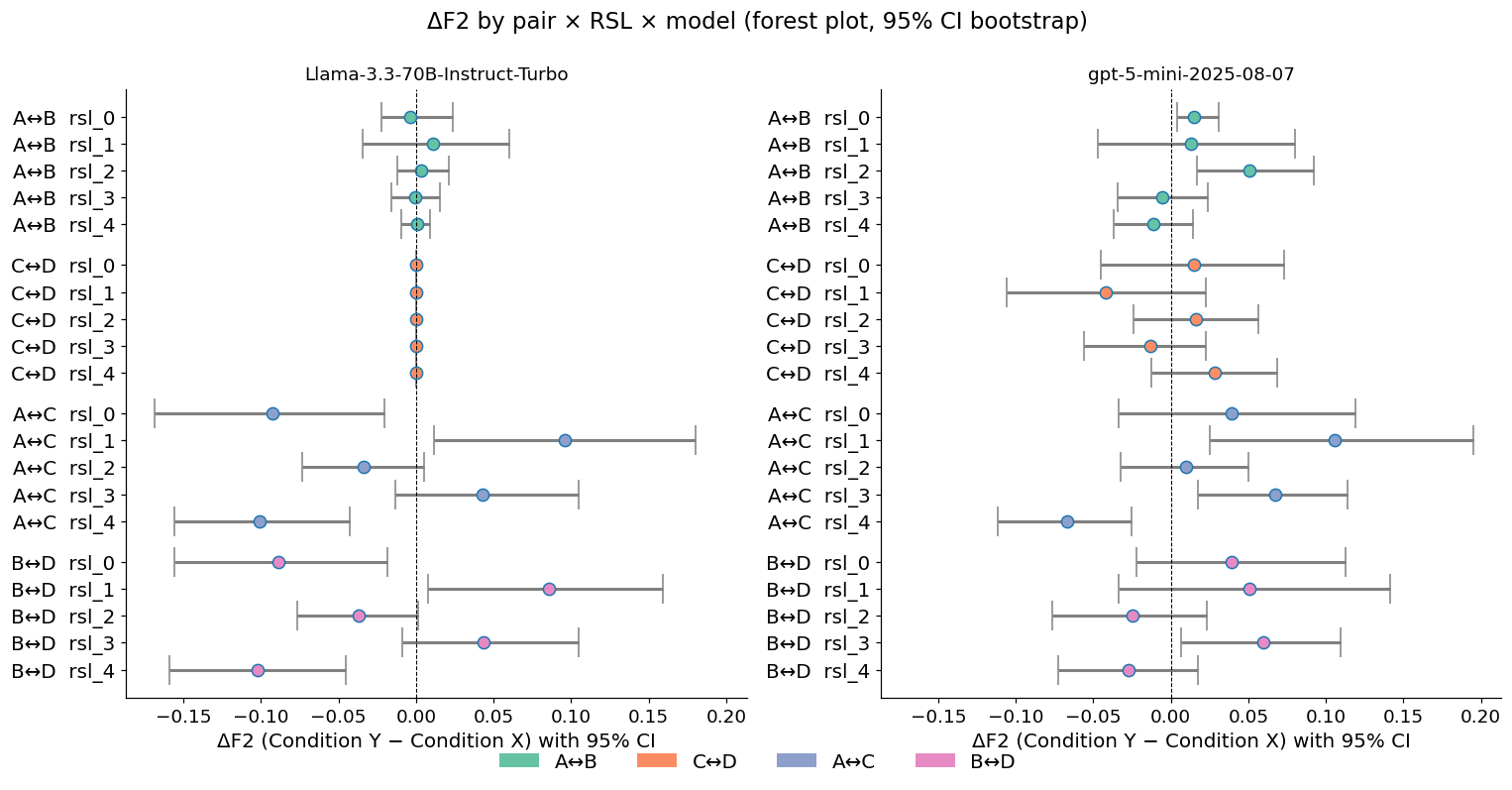}
    \caption{$\Delta$F2 by condition pair, SLR, and model (95\% CI, \textit{bootstrap}). Points to the right of the dashed line indicate F2 gain; points to the left indicate loss.}
    \label{fig:RQ1_forest}
\end{figure}

In individual processing (A$\leftrightarrow$B), the metadata produced a small effect in both models. In Llama, DFR ranged from $0.9\%$ (\texttt{rsl\_1}) to $2.3\%$ (\texttt{rsl\_4}), and the corresponding $\Delta$F2-scores remained close to zero, with 95\% CIs crossing zero in all five reviews (Figure~\ref{fig:RQ1_forest}). In GPT-5-mini, the DFRs ranged from $0.5\%$ to $6.3\%$, with a greater amplitude in \texttt{rsl\_4}. Although McNemar's test indicated statistical evidence in some combinations (
$p < 0.05$ in three of the five SLRs for Llama), the magnitude suggests a small effect. Because models were given the true prevalence rather than a pilot estimate, this result reflects the effect of oracle prevalence metadata, not necessarily the effect of the imperfect estimates available in practice, with no uniform directional pattern of change.

The most notable result occurred in C$\leftrightarrow$D for Llama: DFR was 0.000 in the five SLRs, preserving 6,642 decisions. This suggests that prevalence metadata did not alter decisions when inserted into prompts with multiple studies. The finding is consistent with the hypothesis that, in long contexts, the distribution of attention between items may attenuate contextual information~\cite{liu2024lost}. In addition, the absence of changes indicates low stochastic noise between C and D, which share the same batch prompt structure; this does not rule out variation in A$\leftrightarrow$C and B$\leftrightarrow$D, whose prompts differ structurally, and which would require repeated calls to estimate directly.

In GPT-5-mini, C$\leftrightarrow$D produced DFR between $0.9\%$ and $9.3\%$, with small $\Delta$F2 values (between $-0.042$ and $+0.028$) and 95\% CIs crossing zero in all SLRs. Taken together, metadata, in the tested formulation, do not constitute a stable calibration mechanism: null effect in C$\leftrightarrow$D for Llama and modest and variable effect in GPT-5-mini.

\subsection{RQ2: Does batch processing modify decisions compared to individual processing?}
\label{sec:rq2}

RQ2 examines whether batch processing, in isolation, changes LLM decisions. The comparison A$\leftrightarrow$C isolates this effect, since neither condition includes metadata, while B$\leftrightarrow$D verifies whether the pattern is maintained when metadata are present. In contrast to RQ1, batch processing was associated with substantially larger changes. In Llama, DFR in A$\leftrightarrow$C ranged from $5.6\%$ (rsl\_1) to $26.7\%$ (rsl\_4), with significant McNemar results ($p < 0.001$) in all reviews. The B$\leftrightarrow$D produced a very similar pattern, with DFR from $5.1\%$ to $26.3\%$, which is consistent with C$\leftrightarrow$D = 0 in this model. In GPT-5-mini, the effect was smaller, but still systematic, with DFRs between $1.3\%$ and $14.0\%$ in A$\leftrightarrow$C and significant McNemar results in four of the five SLRs.

The effect of batch processing on performance reveals a relevant contrast between the decision change and the aggregate gain. For Llama in A$\leftrightarrow$C, the sign and magnitude of $\Delta$F2 vary between SLRs (Figure~\ref{fig:RQ1_forest}). There were gains in rsl\_1 ($+0.096$; CI: $+0.012$, $+0.180$) and rsl\_3 ($+0.043$), a value close to zero in rsl\_2 ($-0.034$), and losses in rsl\_0 ($-0.092$) and rsl\_4 ($-0.101$). The case of rsl\_1 is instructive: batch processing increased F2 from 0.37 to 0.46, but 101 studies changed from \textit{Exclude} to \textit{Include}, of which only 29 were correct; at the item level, 29 studies improved and 85 worsened. Similar or opposite patterns in rsl\_3 and rsl\_4 reinforce that interpreting $\Delta$F2 in isolation may hide relevant redistributions of error at the article level.

These pieces of evidence suggest a practical recommendation. Across the five SLRs evaluated, the median $\Delta$F2 of Llama in A$\leftrightarrow$C is close to zero ($-0.034$), and the bootstrap CIs cross zero in two of the five reviews, indicating that, in this set of reviews, batch processing did not show a uniform loss of aggregate performance. Considering that batch processing can substantially reduce inference costs~\cite{pipal2026researchers,cheng2023batch}, it may be an operational option for LLM-assisted screening. However, given the variability of the effect between reviews observed in these results, with $\Delta$F2 ranging from $-0.101$ to $+0.096$ between SLRs, the adoption of batch processing in real pipelines should be accompanied by review-specific validation, ideally with an initial sample executed in individual mode as a reference for comparison.

\subsection{RQ3: How does batch processing, with or without metadata, affect false negatives and direction of decision changes?}
\label{sec:rq3}

RQ3 investigates how batch processing, with or without metadata, affects false negatives and the net direction of decision changes. Since C$\leftrightarrow$D was already discussed in RQ1, the answer comes mainly from the comparison of B$\leftrightarrow$D. In Llama, B$\leftrightarrow$D reproduces A$\leftrightarrow$C almost exactly, because C$\leftrightarrow$D = 0 and A$\leftrightarrow$B is close to zero; therefore, the metadata in Condition B were insufficient to compensate for the batch effect. In GPT-5-mini, B$\leftrightarrow$D presented DFR between $1.1\%$ and $11.0\%$. The sign of $\Delta$F2 remained the same as that of A$\leftrightarrow$C in three SLRs and was reversed in two. In terms of false negatives, reductions were observed in some combinations, such as rsl\_1 and rsl\_3 in both models, and increases in others, such as rsl\_4 in both models. Thus, the combination of batch processing and metadata did not produce a consistent reduction in false negatives: the impact varies by review and by model, and in some cases coexists with an increased risk of losing relevant studies.

The analysis of \textit{the net shift}, that is, the net direction of changes, in A$\leftrightarrow$C reveals a cross-cutting pattern. When SLRs are ordered by increasing prevalence (Table~\ref{tab:netshift}), the four reviews with prevalence $\leq 20\%$ present a positive \textit{net shift}, between $+3.5$ and $+10.5$ percentage points. In these cases, batch processing made Llama more inclusive. In the single almost balanced SLR available (prevalence of $53\%$), \textit{the net shift} was strongly negative (-25.3 points), an observed pattern rather than an established mechanism, given the reliance on a single review at this prevalence level. In GPT-5-mini, the same directional pattern appears in rsl\_4, but there is no directional dominance.

\begin{table}[ht]
\centering
\caption{Net shift of the batch effect without metadata (A$\leftrightarrow$C), by prevalence.}
\label{tab:netshift}
\small
\begin{tabular}{lrrr}
\hline
RSL & Prev. & Llama & GPT-5-mini \\
\hline
rsl\_0 & 2.9\%  & +0.078 & -0.007 \\
rsl\_1 & 3.5\%  & +0.043 & +0.006 \\
rsl\_2 & 10.5\% & +0.035 & -0.003 \\
rsl\_3 & 17.4\% & +0.105 & +0.060 \\
rsl\_4 & 53.0\% & -0.253 & -0.110 \\
\hline
\end{tabular}

\end{table}

This pattern is compatible with a possible tendency toward implicit normalization of decisions within the batch. When processing 20 articles together, the model can approximate its decisions to an intermediate proportion of inclusions, although the experimental design does not allow direct identification of the causal mechanism. This behavior tends to benefit reviews with few true positives because it shifts more decisions from \textit{Exclude} to \textit{Include}, but it may penalize balanced reviews by excessively reducing the inclusion rate. Condition D, even with explicit metadata, did not correct this behavior in the balanced review in either model. The finding is compatible with implicit comparison effects in batch prompting~\cite{pipal2026researchers} and with sensitivity to positioning in long contexts~\cite{liu2024lost}.

\subsection{RQ4: Are the observed effects consistent across models from different families?}
\label{sec:rq4}

RQ4 evaluates whether the effects are sustained across models of different families. Partial consistency is observed between the models. The qualitative pattern is repeated in both cases: the effect of metadata is small, the effect of batch processing is substantially larger, and the direction of changes depends on review prevalence. However, the magnitude of the effects differs. Llama presented DFRs in A$\leftrightarrow$C approximately twice as large as those of GPT-5-mini, with an average of $0.160$ versus $0.057$, in addition to the particular result of C$\leftrightarrow$D = 0. In absolute terms, GPT-5-mini showed more conservative behavior, with a predicted inclusion rate below the true prevalence in most SLRs with prevalence below 20\%, except in the lowest-prevalence review (rsl\_0) and, under batch processing, in rsl\_3. This pattern limits \textit{recall} and is reflected in the absolute F2 lower than Llama in these reviews. In rsl\_2, for example, this operational difference was substantial: Llama in individual mode reached a recall of $0.925$ and F2 of $0.705$, while GPT-5-mini reached a recall of $0.369$ and F2 of $0.409$. Complete precision, recall, F2, FNR, and predicted inclusion rate values for all conditions and reviews are reported in Supplementary Material, Part II~\cite{anonymous_artifacts_2026}.

This contrast corroborates the finding of Huotala et al.~\cite{huotala2025sesr} that variation between reviews may be greater than variation between models, but adds an operational layer. The choice of the most suitable model for assisted screening tends to depend on the profile of the target review. Even when the qualitative pattern converges across models, the magnitude of the differences may still be sufficient to affect practical processing-mode decisions.

\section{Conclusion}

This study investigated how screening metadata and batch processing influence LLM decisions in title and abstract selection. In a factorial design of 2$\times$2 with five SESR-Eval reviews and two models, textual metadata had a limited effect, while batch processing produced larger changes.

Batch processing produced the largest effect, with substantially larger DFRs in the A$\leftrightarrow$C and B$\leftrightarrow$D comparisons. This effect varied according to the review and model profile: in more imbalanced reviews, batch processing tended to make Llama more inclusive, whereas in the single almost balanced review evaluated, it produced more restrictive behavior (an observed pattern, not an established mechanism). Furthermore, aggregate metrics and item-level analyses did not always converge, indicating that gains in the F2-score may coexist with worsening in individual decisions. Thus, batch processing should not be treated only as a cost-reduction strategy; its adoption in LLM-assisted screening should also consider possible changes in decision behavior.

From a methodological perspective, the results reinforce that LLM evaluation in screening should not depend only on global metrics. The combination of aggregate metrics, DFR, \textit{net shift}, and paired article-level analysis proved necessary to characterize how decisions change between experimental conditions.

This study has limitations. The results were obtained from five Software Engineering reviews, with two models, a specific prompt formulation, and a single batch arrangement and execution per condition. In addition, the binary \textit{inclusion}/\textit{exclusion} decision simplifies real screening practice, which can involve uncertainty, discussion among reviewers, and additional contextual criteria.

In future work, we intend to expand the analysis to other datasets, domains (e.g., medical, social science), and model families, as well as to test different batch sizes, position effects within batches, and adaptive strategies that choose between individual and batch processing according to the review profile, complemented by cluster bootstrap resampling for uncertainty estimation and systematic instrumentation of token usage, monetary cost, and execution time.

\section*{Acknowledgments}
The authors thank Cesar School for the institutional and financial support provided through its internal research program, which sustains the AIBL research group to which both authors are affiliated.

\section*{Generative AI use statement}

ChatGPT and Claude were used to support the linguistic revision, textual clarity, editorial refinement, and visual preparation of Figure~\ref{fig:fluxo_metodo}. All scientific content, analyses, interpretations, and conclusions were prepared and reviewed by the authors.

\bibliographystyle{sbc}
\bibliography{references}

\end{document}